\documentclass[sigconf]{acmart}

\usepackage{multirow}
\usepackage{enumitem}
\usepackage{caption}
\usepackage{subcaption}
\usepackage{arydshln}

\AtBeginDocument{%
  \providecommand\BibTeX{{%
    \normalfont B\kern-0.5em{\scshape i\kern-0.25em b}\kern-0.8em\TeX}}}

\newcommand{\ith}{\mbox{${i^{\text{th}}}$}}

\newcommand{\rpm}{\sbox0{$1$}\sbox2{$\scriptstyle\pm$}
  \raise\dimexpr(\ht0-\ht2)/2\relax\box2 }

\newcommand{\removeline}[1]{}
\newcommand{\multilinecomment}[1]{}

\def\Lp{\left(}
\def\Rp{\right)}
\def\LP{\left\{ } 
\def\RP{\right\}}

\copyrightyear{2024}
\acmYear{2024}
\setcopyright{acmlicensed}\acmConference[CODS-COMAD 2024]{7th Joint International Conference on Data Science \& Management of Data (11th ACM IKDD CODS and 29th COMAD)}{January 4--7, 2024}{Bangalore, India}
\acmBooktitle{7th Joint International Conference on Data Science \& Management of Data (11th ACM IKDD CODS and 29th COMAD) (CODS-COMAD 2024), January 4--7, 2024, Bangalore, India}
\acmPrice{15.00}
\acmDOI{10.1145/3632410.3632452}
\acmISBN{979-8-4007-1634-8/24/01}

\begin{document}

\title{BBE-LSWCM: A Bootstrapped Ensemble of Long and Short Window Clickstream Models}

\author{Arnab Chakraborty}
\affiliation{%
  \institution{A2D-AI, Intuit}
  \city{Bangalore}
   \country{India}
  }

\author{Vikas Raturi}
\affiliation{%
  \institution{A2D-AI, Intuit}
  \city{Bangalore}
   \country{India}
  }

\author{Shrutendra Harsola}
\affiliation{%
  \institution{A2D-AI, Intuit}
  \city{Bangalore}
   \country{India}
  }

\renewcommand{\shortauthors}{A. Chakraborty, V. Raturi and S. Harsola}

\begin{abstract}
We consider the problem of developing a clickstream modeling framework for real-time customer event prediction problems in SaaS products like QBO. We develop a low-latency, cost-effective, and robust ensemble architecture (BBE-LSWCM), which combines both aggregated user behavior data  from a longer historical window (e.g., over the last few weeks) as well as user activities over a short window in recent-past (e.g., in the current session). As compared to other baseline approaches, we demonstrate the superior performance of the proposed method for two important real-time event prediction problems: subscription cancellation and intended task detection for QBO subscribers. Finally, we present details of the live deployment and results from online experiments in QBO.
\end{abstract}

\maketitle

\section{Introduction}
\label{sec:into}
\textbf{Objective:} Ability to model clickstream in SaaS products (like QBO) can help improve user experience by making intelligent real-time in-session decisions. Some examples are: 
\begin{itemize}
    \item Subscription cancellation (churn) prediction: Predict in real-time if a user is likely to cancel the product subscription within the next few days. These predictions can be used by CRM teams to design in-session proactive and contextual interventions to retain the user.
    \item Intended task detection: Predict user intent when they land on the help section or chat or dial into support. It can be used to personalize the experience by directly presenting users with relevant knowledge articles / FAQs or routing them to the right agent.
    \item Task abandonment prediction: Forecasting in real-time if a user is likely to abandon the task that they started (like setting up payroll). These predictions can be used to offer intervention and guide them toward completing the task.
\end{itemize}

\textbf{QBO:} We will use QuickBooks Online (QBO) product to demonstrate the application of this framework. QBO is an Intuit cloud-based software product that provides various bookkeeping and accounting-related capabilities to small/medium-size businesses. QBO provides a host of features like issuing invoices, accepting payments, tracking business performance, managing inventory, etc., which makes it challenging to model user behavior.

\textbf{Motivation:} Most studies in clickstream modeling is focused on the eCommerce domain, where the goal is to predict user exits with or without purchase in a session and offer marketing interventions to drive users towards purchase. These models mostly rely on pages visited in the current session to make the predictions. On the other hand, the studies related to the prediction of customer events such as churn do not take the user behavior from the current session into account. There is also a handful of literature on the automated encoding of high-dimensional user-behavior feature interactions for CTR (click-through-rate) prediction, but, how these models perform for real-time customer event predictions using clickstream data as the primary predictor, has not been explored. These approaches are not the most optimal choice for the problems considered here due to the following reasons.
\begin{itemize}
    \item Depending on the event to predict (e.g., churn vs intended task), the importance of current session user behavior vs user activities from beyond the current session changes. A methodology is required that can efficiently incorporate both the current session and beyond-current-session historical user behavior and assign importance automatically based on the underlying data.
    \item In many SaaS products only the pages are not enough to understand the user behavior efficiently and hence, a more generic framework that can incorporate further granular properties of clickstream such as events performed within a page, is required.
    \item To avoid label leakage and to closely replicate the real-time deployment scenario, model training, and evaluation have to be based on reference timestamps (referred to as ref-ts) at which the model inference is requested. There can be multiple ref-ts within a session accumulating a few hundred ref-ts per user making the problem challenging in terms of scale and class imbalance.
\end{itemize}

\textbf{Proposed Approach:} 
In this paper, we present BBE-LSWCM (Block-Bootstrapped Ensemble of Long and Short Window Clickstream Models) -- an ensemble approach for real-time clickstream modeling using both aggregated user behavioral data over a longer historical window (e.g., over the last few weeks) as well as user activities over a short window from recent past (e.g., current session or in last hour). In most of the data systems in the industry, only a few datasets are available in real-time through streaming systems (like Kafka) while most of the other datasets reside in datalake tables and get updated with a frequency of 24 hours or more. Models built on only real-time data are often not robust due to the limited coverage and scope of real-time datasets. On the other hand, predictions coming from models trained only on the daily updated datasets are not sensitive to the dynamic nature of the user behavior. BBE-LSWCM, by its design, can incorporate both the real-time updated and daily updated datasets and model user behavior in real-time robustly. In addition, BBE-LSWCM allows the featurization of the models to be completely automated and data-driven, making this approach reusable for multiple clickstream-based event prediction problems.

\textbf{Results:} 
On the QBO dataset, BBE-LSWCM has achieved at least a 30\% better lift score over the next best model for real-time churn detection. In the online A/B test in QBO, we have observed a 20\% reduction in the first 31 days churn for high-risk customers detected by BBE-LSWCM deployed in a real production setting. Moreover, by rigorous ablation study, the generalisibility of BBE-LSWCM has been proved through application in both churn and intended-task detection problems.

\textbf{Contributions:} 
The main contributions of this paper are:
\begin{itemize}[noitemsep,topsep=0pt]
    \item[(1)] a novel robust ensemble architecture combining long window and short window user behavior for real-time clickstream modeling;
    \item[(2)] block-bootstrap sampling of ref-ts from clickstream for train data generation;
    \item[(3)] generic modeling framework with automated featurization and ability to incorporate multiple granular clickstream properties like page, event, time;
    \item[(4)] a detailed experimental study on QBO dataset;
    \item[(5)] live deployment of BBE-LSWCM to power scalable yet cost-effective in-session proactive interventions.
\end{itemize}

\section{Related Work}
\label{sec:rel_work}
Clickstream modeling has been extensively used in eCommerce domain for multiple problems like next page prediction \cite{bernhard2016clickstream, bogina2017incorporating, gunduz2003web, hidasi2015session, jenkins2019clickgraph}, recommending item to the user \cite{adeniyi2016automated, gu2020hierarchical, li2017neural, quadrana2017personalizing, sun2019bert4rec, you2019hierarchical, zhou2018micro, zhu2017next}
and purchase prediction \cite{nishimura2018latent, vieira2015predicting, yeo2018conversion, ozyurt2022deep}. But as described in Sec~\ref{sec:into}, these approaches do not take the historical user behavior beyond the current session into account while predicting the target. Moreover, most of these studies are restricted to predicting session-related customer events e.g., next page visit or exit without purchase. The applicability of these methods has not been studied for the types of customer events mentioned in Sec.~\ref{sec:into}.

On the contrary, most studies in churn detection \cite{huang2015telco, Ahmad19, Dias2020, Lu02, Ghorbani09, mitrovic2017churn, kdd2021gameschurn} use only the aggregated user-behavior data over a long period of time as input to the model and evaluation is based on a single inference of the model in time. Hence, the methods and results of these papers can not be generalized for real-time customer event prediction problems. In \cite{yang2018know}, though the authors have proposed a churn modeling framework that considers a sequence of daily activity data as a multivariate time-series input, this methodology does not take the in-session user behavior or real-time deployment scenario into account. There are few recent studies \cite{bertens2017games, deligiannis2020designing} which propose a framework for real-time churn identification, but these studies do not discuss automated use of large-scale clickstream data and application of these methods for other types of customer events such as intended task detection is not straightforward.

For problems such as intended task detection or task abandonment prediction using user behavior data, models introduced for CTR prediction in Ads recommendation \cite{cheng2016wide, guo2017deepfm, lian2018xdeepfm, chen2021enhancing, zhou2018deep} can also be used. Though these neural network-based architectures provide an automated way to efficiently capture the high-dimensional cross-interactions between the numerical and categorical user behavior features, the efficacy of these models to capture the intrinsic sequential nature of the high-dimensional clickstream data, for the customer event prediction problems considered in this paper, have hardly been explored in literature.

To the best of our knowledge, there is no prior work that has addressed all the practical industry challenges mentioned so far in real-time modeling of customer events using clickstream data in the SaaS domain.

\section{BBE-LSWCM: Block-bootstrapped Ensemble of Long and Short Window Clickstream Models}
\label{sec:method}
\subsection{Inference Mechanism}
\label{subsec:inf_mech}
We consider the following inference task: given a ref-ts $T$, on request from the model consumer (product UI, customer success tool, etc.), the concerned task is to predict the \textit{target} with real-time latency. \textit{Target} could be the risk of subscription cancellation within a few days from $T$, or abandonment of the current task within a short time window from $T$, or the intended task of the user at time $T$.

\textbf{Output:} For a ref-ts $T$, for $\ith$ user, the output of this prediction task is a time-dependent variable $y_i(T)$ referred to as \textit{target}. We denote the set of eligible ref-ts's for the $\ith$ user, as $\mathbb{T}_i$. For example, in the real-time churn detection problem, $y_i(T) = 1$ if the $\ith$ user will cancel their subscription during the time-period $(T, T + \delta]$, where typically $\delta > 0$ is 24 or 48 hours. Here, $\mathbb{T}_i$ can consist of all the time-points $T$ at which the $\ith$ user has an active subscription, and the model consumer can request an inference. On the other hand, for the problem of intended task detection of the user, $y_i(T)$ is a multi-class random variable with $\mathbb{T}_i$ containing all the timestamps when the user engages with a help channel for the first time, e.g., the timestamp corresponding to the click to open the digital assistant, or starting the search or, the timestamp when the user calls into the customer care. Note that, unlike the previous works in the eCommerce domain on clickstream modeling \cite{ding2015learning, hatt2020early, ozyurt2022deep} where the target is always defined at the end of the clickstream session and only single inference within a session is considered, the formulation considered in this paper supports multiple inferences of the model even within a session and the target can occur further away from the end of a session as well.

\textbf{Input:} Input to this prediction task is the historical clickstream data and user profile information up to time-point $T$. Often in large SaaS products like QBO, the user can perform multiple types of activities such as invoicing, banking, payments, payroll, etc. and so, the start and end of a session are not always well-defined. Hence, we define the clickstream data based on a time-window $[T - w, T)$ for a given ref-ts $T \in \mathbb{T}_i$ and a time duration $w > 0$. The clickstream data for $\ith$ user is $\mathbb{C}_i(T;w) = \langle (t_i^m, p_i^m, e_i^m)\rangle_{m = 1}^{M_i(T; w)}$, with $p_i^m \in \mathbb{P} = \LP P_1, P_2, \dots \RP$ being the page visit and $e^m_i \in \mathbb{E} = \LP 
E_1, E_2, \dots \RP$ is the event (action) performed at timestamp $t_i^m$ and $\langle  \cdot \rangle$ denotes an ordered sequence of objects. To give an example, on \textit{`banking'} page, the user can perform events like \textit{`bankconnect: started', `account: viewed', `transaction: viewed'} etc. The total number of clicks performed by the $\ith$ user in $[T - w, T)$ is $M_i(T; w)$ and $T-w < t_i^1 < ... < t_i^{M_i(T; w)} <= T$ are the ordered timestamps at which the clicks are performed by the user. For simplicity of notation, we drop subscript $i$ from $\mathbb{C}_i(T;w)$ and other variables, when referring to a generic user.

User profile information is also another input to the model. For $\ith$ user the user profile information at time $T$ is denoted by $z_i(T)$ which is a multivariate variable consisting of information such as the subscription type, tenure duration, location, and any other demographic information that is available.

\subsection{Model Architecture for BBE-LSWCM}
\label{subsec:model_archi}
The proposed BBE-LSWCM approach is a hybrid ML framework tailored to solve real-time event prediction tasks. It allows the prediction to incorporate not only the `recent' in-session user activities but also, the historical user behavior from a longer window.
\begin{figure}[h]
  \centering
  \includegraphics[trim={0.05cm 0.1cm 0.1cm 0.1cm}, width=.98\columnwidth, height=0.2\textheight, scale=0.9]{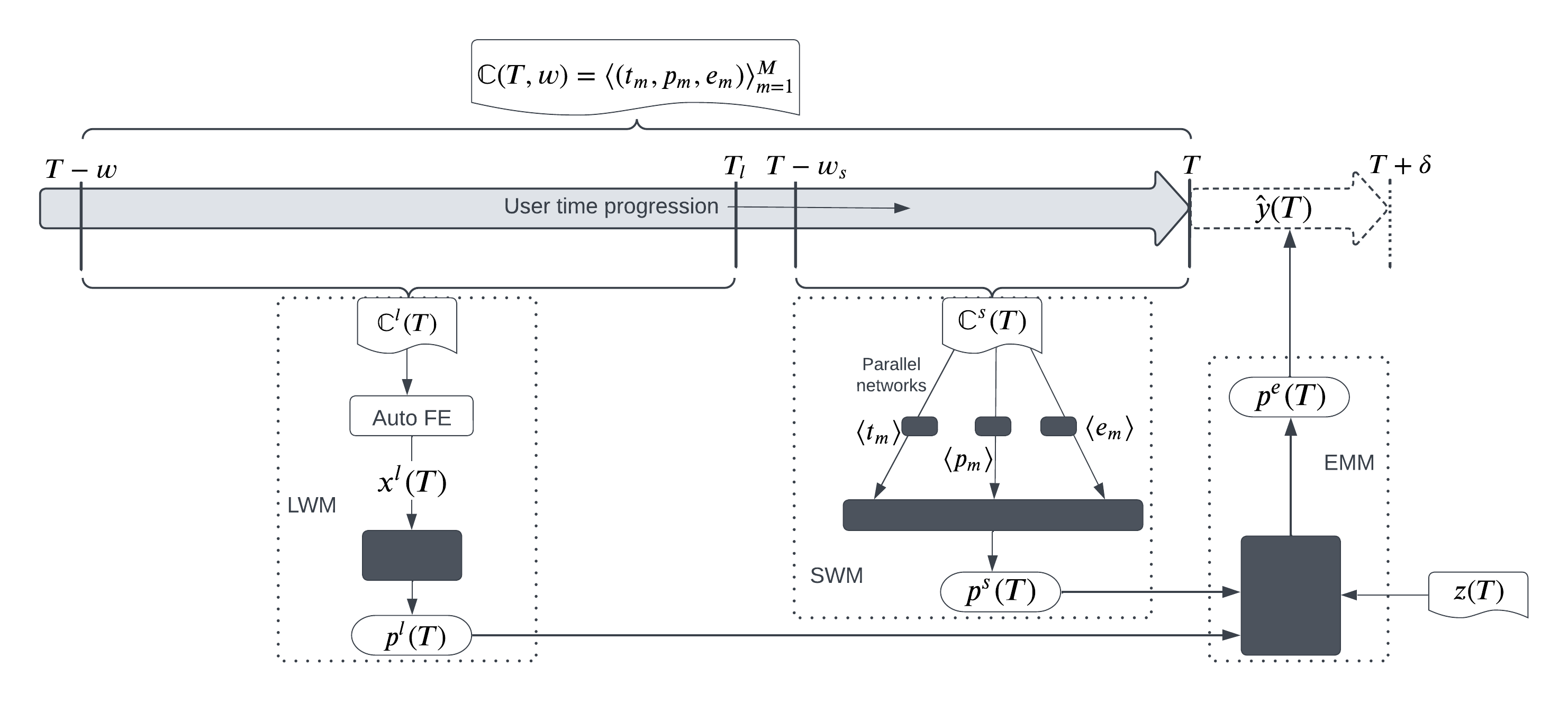}
  \caption{\small BBE-LSWCM inference mechanism (black rectangles denote ML modules).}\vspace{-1mm}
    \label{fig:inf_mech}
\end{figure}
As mentioned earlier in Sec.~\ref{sec:into}, in SaaS subscription products such as QBO, often the user behaviors beyond the current session influence the distribution of the target. For example, to predict whether a customer will cancel their subscription within the next few hours or days, in addition to the in-session clickstream, historical user behavior such as activities over the last few weeks in some of the important pages is highly important. On the other hand, for detecting the user's intended task when they are engaging with help channels for the first time, often the in-session user behavior is the key predictor. Hence, depending on the data and problem in hand the modeling framework needs to automatically calibrate the importance of in-session vs longer history of user behavior. Moreover, for a given user, $\mathbb{C}\Lp T; w \Rp$ can contain more than $10^5$ clicks if $w$ is a few weeks. Processing such a huge volume of clicks in real-time for model inference would be computationally challenging resulting in high inference latency and would also be highly expensive in terms of deployment cost.

As shown in Fig.~\ref{fig:inf_mech}, in BBE-LSWCM, we propose to break the historical clickstream $\mathbb{C}\Lp T; w \Rp$ into two parts that are mutually exclusive in time. One part contains the historical clicks that happened in the same session as of the ref-ts $T$ or in the very recent past (30 minutes to one hour back) from $T$. We call this clickstream as \textit{Short Window Clickstream (SWC)} and denote it by $\mathbb{C}\Lp T, w_s \Rp$ (for simplicity denoted as $\mathbb{C}^s\Lp T \Rp$), where $w_s > 0$ can usually vary up to a few hours. The second part of the clickstream, named as \textit{Long Window Clickstream (LWC)}, consists of the clicks that happened in the longer historical time window $(T - w, T_l]$ for a time point $T_l$ such that $T - w << T_l < T-w_s$. So, this clickstream data does not contain the clicks from recent history, rather it contains clicks from a further past. We denote LWC as $\mathbb{C}\Lp T_l; w_l \Rp$ (for simplicity denoted as $\mathbb{C}^l\Lp T \Rp$), where $w_l = w - (T - T_l)$ is the length of the long historical window. Instead of building a single model with $\mathbb{C}\Lp T; w \Rp$ as an input -- which requires processing thousands of clicks at the time of every real-time inference -- we propose a step-wise modeling approach by focusing separately on the recent and further past user activities. We showcase in the later sections how this approach can achieve not only lower deployment latency and cost but also better predictive accuracy.

BBE-LSWCM has 3 major components. The (1) \textbf{Long Window Model (LWM)} estimates the conditional probability of the target given the user's clickstream activities from the long historical window, i.e., $P(y(T) | \mathbb{C}^l(T))$. The output of LWM is denoted by $p^{l}(T)$. We make the inference of LWM as a (daily) batch job and make the prediction of LWM available as an input to the real-time inference. By doing this, we reduce the latency and cost of real-time inference significantly while not losing the user behavior context coming from the long historical window. The (2) \textbf{Short Window Model (SWM)} estimates the conditional distribution of the target $P(y(T) | \mathbb{C}^s(T))$ given the user's recent in-session clickstream. The output of the SWM is denoted by $p^{s}(T)$. This model is evaluated in real-time by allowing the prediction to be sensitive to the user's actions in the recent past (in-session). Finally, (3) \textbf{Ensemble Meta Model (EMM)} takes $p^{s}(T)$, $p^{l}(T)$ and the user profile information $z(T)$ as inputs to finally estimate the conditional distribution of the target $P(y(T) | \mathbb{C}(T; w), z(T))$. The output of EMM is denoted by $p^e(T)$ and appropriate thresholding is applied on $p^e(T)$ to finally get the prediction $\hat{y}(T)$. Like SWM, the EMM is also evaluated at the time of real-time inference. In the next sections, we showcase the advantages of this approach: (a) because of the proposed block-bootstrapped ensemble and profile-likelihood-based estimation, BBE-LSWCM has better performance both in terms of prediction accuracy and early warning generation, and (b) the latency and implementation cost of BBE-LSWCM is significantly lower as compared to a joint model which uses a long window of historical clickstream (more than few weeks) to predict the target at the time of real-time inference.\vspace{-3mm}

\subsection{Model Components}
\label{subsubsec:model_comp}
\textbf{LWM:} The length of the LWC, $\mathbb{C}^l(T) = \langle (t^m, p^m, e^m)\rangle_{m = 1}^{M(T_l; w_l)}$, i.e. $M(T_l; w_l)$ (for simplicity denoted as $M^l(T)$), can be as large as $\sim 10^5$. To process such a large volume of varying length clickstream we propose an automated feature engineering block (\textit{Auto FE} in Fig.~\ref{fig:inf_mech}) in BBE-LSWCM. For inference at given ref-ts $T$, we define the input feature vector of LWM, $x^l(T)$, as:
\begin{equation}
\label{eq:lwm_feat}
\begin{split}
C(P_j) &= \frac{1}{\tau(T)}\sum_{m=1}^{M^l(T)} \mathbb{I}\Lp p^m = P_j \Rp, C^{P} = \frac{1}{\tau(T)}\sum_j\sum_{m=1}^{M^l(T)} \mathbb{I}\Lp p^m = P_j \Rp,
 \\
x^{l}(T) &= [ \Lp C(P_j) \Rp_{j=1}^{\lvert \mathbb{P} \rvert}, \Lp C(E_k) \Rp_{k=1}^{\lvert \mathbb{E} \rvert}, \Lp C^{P}, C^{E}\Rp ],
\end{split}
\end{equation}
where $\tau(T)$ is the duration the user is using the product for at $T$, $[ \; ]$ denotes the concatenation of the vectors, $\lvert \cdot \rvert$ denotes the size of a set, and, $C(E_k), C^E$ is similarly defined as of $C(P_j), C^P$. The above featurization creates an \textit{additive-count-vectorized} feature vector $x^{l}(T)$ of size $(\lvert \mathbb{P} \rvert+\lvert \mathbb{E} \rvert+2)$ to measure the user activities over the long historical window. Note that, this featurization can be even further extended to include more properties of clickstream beyond page and event, e.g. type of the action (scroll, swipe, click), etc.

We model the conditional distribution of the target $y(T)$ non-parametrically
\begin{equation}
\label{eq:lwm_est_form}
\begin{split}
P\Lp y(T) | \mathbb{C}^l(T) \Rp = \phi^{l} \Lp x^{l}(T); \Theta^{l} \Rp  
 = \sum_{a=1}^A f_a \Lp  x^{l}(T); \theta_a^{l} \Rp,
\end{split}
\end{equation} with tree-based weak learners $f_a\Lp \cdot ; \theta_a^{l}\Rp$ with unknown parameters $\Theta^{l} = \Lp\theta_a^{l}\Rp_{a=1}^A$. The details of the estimation are mentioned in Sec.~\ref{subsec:param_est}. The output of LWM is denoted by $p^{l}(T) = \phi^{l} \Lp x^l(T); \hat{\Theta}^{l} \Rp$, where $\hat{\Theta}^{l}$ is the estimated parameter values.

\textbf{SWM:} In SWM, we separate out the input clickstream $\mathbb{C}^s(T)$ of length $M(T; w_s)$ ($ = M^s(T)$) to three sequences of timestamps, pages, and events. We convert the page and event sequences into a sequence of tokens coming from vocabularies $\mathbb{P}, \mathbb{E}$ respectively. The input to SWM is $x^s(T) = [ \langle o(p^m) \rangle_m, \langle o(e^m) \rangle_m, \langle t^m \rangle_m]$ with $o(p)$ is the one-hot-encoded representation of $p \in \mathbb{P}$ and same for the events. For large online subscription products like QBO, $ \lvert \mathbb{P} \rvert$ can be more than $10^2$ and $\lvert \mathbb{E} \rvert$ can be as large as  $10^4$. Hence, to reduce the dimension, we convert the one-hot-encoded representation to a fixed-length embedding vector as
$r_p(T) =  \langle o(p^m) \rangle_m \cdot R_p$ where $R_p$ is the $M^s(T) \times B$ embedding matrix such that $B << \lvert \mathbb{P} \rvert$. Similarly, we have an embedding matrix $R_e$ for the events. To learn the sequential context information from page-event sequences, instead of random initialization, we propose to initialize $R_e$ and $R_p$ by training an unsupervised Skip-Gram model and then allow the elements of these matrices to be updated through back-propagation while training for downstream tasks.

To model the intrinsic sequential nature of the in-session clicks, we use parallel Bidirectional LSTM (BiLSTM) Layers for pages and events as follows,
\begin{equation}
\label{eq:bilstm}
    h_p^m = LSTM\Lp h^{m-1}_F, r(p^m) , W_F , b_F \Rp \oplus LSTM\Lp h^{m+1}_B, r(p^m) , W_B , b_B \Rp,
\end{equation} where $r(p^m)$ is the embedding representation of the $m$-th visited page and $\oplus$ denotes the element-wise addition. A similar computation is done in the parallel events network (to get $h_e^m$). As user clicks are often not uni-directional in nature, BiLSTM helps to extract additional context information from the user clicks. We stack multiple BiLSTM layers to extract important sequential information from page and event sequences and, then to summarize the most important features over time, Global Max Pooling (GMP) and Global Average Pooling (GAP) are performed on the final sequential hidden layers $\mathbf{h}_p = [h_p^1, \dots, h_p^M]$ and $\mathbf{h}_e = [h_e^1, \dots, h_e^M]$. For the timestamp sequence $\langle t_m \rangle_m$, we pass it through a feed-forward-neural network (FNN) and concatenate the output of FNN ($\Tilde{t}$) with the global pool layers of page and event networks as shown below.
\begin{equation}
\label{eq:swm_pred_layer}
\begin{split}
&\tilde{h}^M = [\text{GMP}\Lp \mathbf{h}_p \Rp, \text{GAP}\Lp \mathbf{h}_p \Rp, \text{GMP}\Lp \mathbf{h}_e \Rp, \text{GAP}\Lp \mathbf{h}_e \Rp, \Tilde{t}],\\
&h_1 = \text{Leaky-RELU} \Lp W_1 \cdot \tilde{h}^M + b_1 \Rp, h_2 = \text{Leaky-RELU} \Lp W_2 \cdot h_1 + b_2 \Rp.
\end{split}
\end{equation} Finally, as shown above, a two-layer FNN in conjunction with a softmax output layer is used to get the output of SWM, i.e., $p^{(s)}(T) = \text{softmax}\Lp W_3 h_2 + b_3 \Rp$ (see Eq.~\eqref{eq:swm_pred_layer}). Let us summarize the SWM network in Eq.~\eqref{eq:bilstm}-\eqref{eq:swm_pred_layer} as $p_i^{s}(T) = \phi^{s} \Lp x_i^s(T); \Theta^{s} \Rp$, with $\Theta^{s}$ denoting all the parameters in SWM network.

\textbf{EMM:} For EMM, we construct the input feature vector $x^{e}(T)$ as,
\begin{equation}
\label{eq:emm_input}
\begin{split}
x^{e}(T) = [ p^{l}(T), p^{s}(T), z(T) ], 
\end{split}
\end{equation} where, $z(T)$ is the dynamic user-profile information and $p^{l}(T), p^{s}(T)$ are the output vectors from LWM and SWM respectively with size $C-1$ where $C$ is the number of possible classes for the target $y(T)$. Finally, the conditional distribution of the target given the entire history of clickstream data, starting from a long window (few weeks in history) to the in-session clicks, and the user profile information is modeled as,
\begin{equation}
\label{eq:emm_model}
\begin{split}
P(y(T) | \mathbb{C}(T; w), z(T))  & \approx P(y(T) | \mathbb{C}^{l}(T), \mathbb{C}^{s}(T), z(T))\\
&= \phi^{(e)} \Lp x^{e}(T); \Theta^{e} \Rp.
\end{split}
\end{equation} In BBE-LSWCM we allow the $\phi^{(e)}\Lp \cdot \Rp$ to be modeled as a regularized logistic regression or tree-based ensemble model depending on the prediction task.

\subsection{Parameter Estimation}
\label{subsec:param_est}
The joint formulation of BBE-LSWCM can be written as (Eq.~\eqref{eq:emm_model}),
\begin{equation}
\label{eq:joint_param}
\begin{split}
P(y_i(T) | \mathbb{C}_i(T; w), z_i(T)) = \phi^{e} ( [& \phi^l( x_i^l(T); \Theta^{l}), \phi^l\Lp x_i^s(T); \Theta^{s}\Rp,\\
& z_i(T)  ]; \Theta^{e} ).
\end{split}
\end{equation} To estimate all the model parameters $\Theta = [\Theta^{l}, \Theta^{s}, \Theta^{e}]$ we have historically observed training data $$\mathbb{D} = \LP \Lp y_i(T), \mathbb{C}_i(T, w) \Rp : 1 <= i <= N, T \in \mathbb{T}_i \RP,$$ with $N$ being the number of unique users and $\mathbb{T}_i$ is the set of `eligible' reference time-points in the training data for the $\ith$ user. Note that, churn and task abandonment type customer events are usually rare events making the training data imbalanced in nature. Moreover, for problems like real-time churn detection, only the last few observations for churned users are from the positive class amplifying the class imbalance even further. Also, including one training record corresponding to every eligible ref-ts $\mathbb{T}_i$ will result in a data explosion. To handle these challenges, we propose a Block-Bootstrap-based sample data selection and training of different components of BBE-LSWCM on independent bootstrap samples.

\textbf{Block Bootstrap Sampling (BBS):} In BBE-LSWCM parameter estimation, we use BBS to generate a bootstrap sub-sample of reference time points denoted by $\mathbb{T}^b \subset \mathbb{T} = \LP T_1 < ... < T_n \RP$ for each user. To do so, first the reference time points in $\mathbb{T}$ are grouped into blocks as $\mathscr{B}_k(L) = \LP T_j : j = (k-1)L + l, 1 \leq l \leq L\RP$ with $L > 0$ being the block length (or, block size). Next, let us consider the collection of such non-overlapping blocks $\mathbb{B} = \LP 
\mathscr{B}_k(L) : 1 \leq k \leq \lfloor n/L \rfloor \RP$. We randomly sample $K < \lfloor n/L \rfloor$-many blocks from $\mathbb{B}$ without replacement and collate the ref-ts's in the sampled blocks to generate $\mathbb{T}^b$. This block-wise bootstrap technique is used in time series and other dependant data applications to simulate bootstrapped samples for estimation of higher order statistic \cite{lahiri1999theoretical, Lahiri03}. By allowing consecutive reference time points to be considered in the sampled training date, BBS brings certain robustification which is demonstrated in the additional experimental results. For problems with high class imbalance, we only down-sample from the majority class through BBS to increase the minority-to-majority class ratio in training data.

\textbf{BBE-LSWCM parameter estimation:} For the estimation of BBE-LSWCM parameters $\Theta$, instead of joint estimation using the full training data $\mathbb{D}$, we propose step-wise estimation of LWM, SWM, and EMM parameters as described in Algorithm~\ref{alg:param_est}.
\newfloat{algorithm}{t}{lop}
\begin{algorithm}
\caption{Estimation of BBE-LSWCM Parameters}\label{alg:param_est}
\begin{itemize}
    \item[1.] LWM Parameter Estimation:
    \begin{enumerate}
        \item[I.] Use simple random sample without replacement to obtain a sample set of users $\mathbb{N}^l = \LP i_1, \dots, i_{N^l} \RP \subset \LP 1, \dots, N \RP = \mathbb{N}$. For each user $i \in \mathbb{N}^l$, use BBS to select a random sample of reference time points $\mathbb{T}_i^l \subset \mathbb{T}_i$.
        \item[II.] Use LightGBM \cite{ke2017lightgbm} to estimate the LWM parameters by minimizing the cross-entropy loss over the bootstrap sample for LWM training $\mathbb{D}_l = \LP \Lp y_i(T), x^l_i(T) \Rp : T \in \mathbb{T}_i^l, i \in \mathbb{N}^l \RP$ \begin{equation}
        \label{eq:lwm_est_param}
        \begin{split}
 \hat{\Theta}^{l} &= \underset{\Theta}{\text{argmin}} 
 \; -\sum_{\mathbb{D}^l} \sum_{c=1}^C \mathbb{I}\Lp y_i(T) = c \Rp \log \Lp \phi^{l} \Lp x^{l}_i(T); \Theta \Rp \Rp,
\end{split}
\end{equation} where $\sum_{\mathbb{D}^l}$ denotes the summation over the observations in $\mathbb{D}_l$.
    \end{enumerate}
    \item[2.] SWM Parameter Estimation:
    \begin{enumerate}
        \item[I.] Repeat step 1.I to generate another independent bootstrap training sample for SWM $\mathbb{D}^s = \LP \Lp y_i(T), x^s_i(T) \Rp : T \in \mathbb{T}_i^s, i \in \mathbb{N}^s \RP$.
        \item[II.] Use ADAM-based gradient descent algorithm to estimate the SWM network parameters by minimizing the cross-entropy loss over $\Lp \mathbb{D}^{s} \Rp$ similar to Eq.~\eqref{eq:lwm_est_param}.
    \end{enumerate}
    \item[3.] EMM Parameter Estimation:
    \begin{enumerate}
        \item[I.] Repeat step 1.I to generate a bootstrap training sample $\mathbb{D}^e = \LP \Lp y_i(T), \mathbb{C}_i(T), z_i(T) \Rp : T \in \mathbb{T}_i^e, i \in \mathbb{N}^e \RP$.
        \item[II.] Compute the EMM input features $x_i^e(T)$ using Eq.~\eqref{eq:emm_input} where $p_i^l(T) = \phi^l\Lp x_i^l(T), \hat{\Theta}_l \Rp, \; p_i^s(T) = \phi^s\Lp x_i^s(T), \hat{\Theta}_s \Rp.$ Estimate the EMM parameter $\Theta^e$ by minimizing the cross-entropy loss over the EMM training data $\mathbb{D}^e$.
    \end{enumerate}
\end{itemize}\vspace{-2mm}
\end{algorithm} By taking independent bootstrapped samples from $\mathbb{D}$ for separately estimating LWM, SWM, and EMM parameters, we guard against the overfitting problem in stacked-ensemble approaches. Moreover, the estimation procedure in Algorithm~\ref{alg:param_est} uses profile-likelihood-based estimation \cite{severini1992profile, murphy2000profile} of BBE-LSWCM parameters. To estimate the EMM parameters $\Theta^e$ in Eq.~\eqref{eq:joint_param}, the LWM and SWM parameters ($\Theta^l$, $\Theta^s$) behave like `nuisance' parameters and hence, these are first estimated separately from bootstrapped samples and then $\Theta^e$ is estimated by profiling out $\Theta^l$ and $\Theta^s$. Asymptotically the estimated parameters through profile-likelihood methods converge to the true values \cite{murphy2000profile}.

\section{Experimentation Results}
\label{sec:exp_result}
We have chosen two important problems related to QBO to demonstrate the merits of the BBE-LSWCM for real-time customer event predictions using clickstream. QBO is used by millions of small businesses to manage their financial accounts and hence, our experimental data set is highly dynamic in nature and includes a large variety of clickstream behavior and user personas. In the first problem, the prediction task is to infer whether a user is at high risk of canceling their subscription within the next 48 hours (look-ahead window chosen based on CRM team input). This is a binary classification problem and we will refer to this problem as \textit{real-time churn prediction}. The other problem referred to as \textit{intended task detection}, requires predicting, in real-time, the task the user is concerned about as they engage with the help channels, e.g., banking, payroll, tax, etc. This is a multi-class classification problem.

\subsection{Test Data}
\label{subsec:test_data}
\textbf{Real-time churn prediction:} In this problem we focus on QBO subscribers who are First Time Users (FTU) i.e., the customers who are in their first billing cycle and hence, are more inclined to cancel the subscription because of difficulty and ambiguity in using the product at its full potential and lack of support \cite{yang2018know}. Based on marketing research, the efficacy of static offline retention campaigns such as email, message, or phone is lower for these customers \cite{van2012online, retana2016proactive}. Hence, an in-session intervention powered by a real-time churn prediction system is necessary to reduce the churn rate for FTU. For the results reported in this paper, we have curated a sample dataset of 50,000 FTU subscribers over a period of 8 months. Sample data was curated to keep the proportion of churned customers similar to other studies in literature \cite{Ghorbani09, Burez09, mitrovic2017churn}, at 8\%. We take the first 6 months as `train period' for training and the last two months are used as test duration for reporting the evaluation metrics. To replicate the real-time deployment scenario (see Sec~\ref{subsec:eval_cri}), we evaluate the model multiple times within a day for each user when they are online in the product. Thus, the number of combinations of user and ref-ts in the training is around 2 million, and for the test is around 700 thousand.

\textbf{Task-intent detection:} For this problem, the target is a multi-class random variable with 20 possible classes, e.g., [banking, chart, estimates,...]. The first timestamp where the user has interacted with any of the help channels (digital assistant, search, or call to customer care) is considered as the ref-ts ($T$), and the corresponding task-intent recorded by the channels post interaction is used to train and evaluate the model. The data considered for training has around 500,000 records of user and ref-ts combinations with their intended tasks and the same for testing is approximately 200,000. The class distribution is highly skewed with the top three most frequent categories contributing more than 40\% of all the records.

\textbf{Clickstream and user-profile data:} We consider all the timestamped clicks (timestamp - page - event) of the users for the above-mentioned time frame of 8 months as input to the model. Based on the dataset considered here, on average in a day the number of clicks per active user can vary between 50 to 5000, and, the entire clickstream data considered in this paper contains over 1 billion clicks with around 350 unique pages and 2500 unique events, making this data-set very diverse. In addition to the clickstream data, we also consider several dynamic user-profile information such as the duration of the user in the product from the start of the subscription in days, the type of the subscription, type of onboarding, type of industry, etc. as input to the models.

\subsection{Evaluation Criterion}
\label{subsec:eval_cri}
In the previous studies on churn detection, mostly the evaluation is done based on a single inference for each user and using usual binary classification metrics such as AUROC, accuracy, and lift scores at certain deciles. In some of the recent research in the eCommerce domain to predict 'customer exits without purchase' using clickstream analytics \cite{hatt2020early, ozyurt2022deep}, to prove the efficacy of the models for early warning generations, the model is inferred at pages which are few steps earlier to the user exit, but still a single inference is used to measure the accuracy of the model. Moreover, this approach of evaluation is not reflective of the real-time deployment setup, since it requires the model trigger mechanism to know about the number of steps to user exit beforehand, which is not possible.

In this paper, to make the evaluation set-up representative of the production use case, we generate the model inference for each eligible user (FTU and not churned yet) multiple times within a day when they were online in the product. Let us denote all the predictions (churn risk or propensity) for $\ith$ user during the test period as $\LP p_i(T_j) : T_j \in \mathbb{T}_i \RP$. To make our results comparable to previous studies related to customer churn and, to meet marketing requirements of intervening a certain percentage of users based on predicted churn risk, we propose aggregating the predictions as $\hat{y}_i = \underset{T_j \in \mathbb{T}_i}{\text{max}} \; \mathbb{I} \LP p_i(T_j) > \lambda_d \RP$. Here, $\lambda_d$ is a threshold chosen from the validation data in such a way that $\sum_i\hat{y}_i$ is approximately equal to $(10\times d )\%$ ($d$-th decile) of the total number of unique users. Note that, $\hat{y}_i = 1$ implies that the model has at least once identified the $\ith$ user as at high risk of churn during the considered period. Finally, we compare the aggregated predictions $\hat{y}_i$ with user-level churner vs non-churner, i.e. $y_i = \underset{T}{\text{max}} \; y_i(T)$, to compute the following two evaluation metrics.
\begin{equation}
\label{eq:lift_score}
\begin{split}
 \text{DL$d$} = \frac{10 \times \sum_i \Lp y_i \times \hat{y}_i \Rp}{d \times \sum_i y_i},\;\; \text{ATWC$d$} &= \frac{\sum_i (\Tilde{T}_i - \hat{T}_i ) \times \Lp y_i \times \hat{y}_i \Rp}{\sum_i   \Lp y_i  \times \hat{y}_i \Rp },
\end{split}
\end{equation} with $\Tilde{T}_i = \text{max}\LP \mathbb{T}_i \RP$ is the time the $\ith$ user has churned (if churner) and $\hat{T}_i$ denotes the first time the model has predicted high risk of churn of this user, i.e., $\hat{T}_i = \text{min} \LP T_j : p_i(T_j) > \lambda_d, T_j \in \mathbb{T}_i \RP$. DL$d$ is the decile lift at decile $d$ and ATWC$d$ is the average time from warning to churn. From the definitions in Eq.~\eqref{eq:lift_score}, for a given decile $d$, DL$d$ quantifies the lift in the churn identification as compared to a non-predictive random model and ATWC$d$ measures the average time from the first detection to churn for the true-positive cases. Additionally, we also evaluate the models w.r.t. the AUROC score at a user level to evaluate the classification ability of the models.

For the intended task detection, we compare the models in terms of one-vs-rest AUROC score (ovr-AUROC) \cite{ovrroc2022}, as this is a multi-class classification problem. As the dataset is highly imbalanced in terms of class occurrences and there is no proper business insight on which task to give more importance to, we have used macro-averaging of the class-specific OvR-AUROC scores.\vspace{-2mm}

\subsection{Baseline Approaches}
\label{subsec:baseline_approaches}
\textbf{Baselines:}
For the real-time churn detection problem, we have compared BBE-LSWCM with a set of baselines which can be categorized into the following four buckets. (1) \textbf{Batch models:} The most commonly used churn detection frameworks use batch predictions, where the propensity of churn is generated for eligible customers on a weekly or monthly basis and the marketers use those for driving static retention campaigns. This model uses domain information to hand-code the features before passing them to the model. Inspired by earlier studies \cite{yang2018know, bertens2017games}, we have considered Logistic Regression (LR), Gradient Boosting Trees (GBDT), and Random Forest (RF). In addition to this, we have also considered the PLSTM architecture \cite{yang2018know} using activity embedding (PLSTM) where we provide the daily activity sequence over the last few days as a multivariate time series input to the model. Similar to previous works \cite{vafeiadis2015comparison, yang2018know, Ghorbani09}, we have used a fixed length (size 30) hand-coded feature vector from domain knowledge which includes user profile information such as type of subscription, onboarding channels, age of the subscription etc. and user behavior features such as number of clicks, number of page visits, time spent etc. in some of the most important product features in QBO. (2) \textbf{In-session models:} This group of models takes the in-session clickstream and user profile information as input to predict the target. Within this group, we have considered a Gradient Boosted Tree model with count-vectorized clicks and user-profile features (GBDT); a multi-layer LSTM model with one-hot-encoded pages as input sequence combined with user-profile features at the flattened layer (LSTM); multi-layer parallel LSTMs with an embedding layer for page sequences and timestamp sequences (emb-LSTM). (3) \textbf{Survival models:} Survival models are also used for some of the user exit detection problems by modeling the time to exit \cite{xu2022lads, Martinsson16}. We have included a batch model with Random Survival Forest (batch-RSF) \cite{ishwaran2008random} and an in-session model DeepSurv (in-ses-DeepSurv) \cite{katzman2018deepsurv} in our set of baselines. (4) \textbf{Joint models:} Like BBE-lSWCM, this set of modeling architectures supports both historical user behavior features and in-session clickstream as inputs along with user profile information. We have considered xDeepFM \cite{lian2018xdeepfm} and DIN \cite{zhou2018deep} from CTR prediction literature where the sequence of page visits is passed as the variable length user-behavior feature and user-profile data is passed as fixed length feature vectors. We also consider CDMM \cite{ozyurt2022deep} from the user exit prediction paradigm using clickstream data. As input, this model takes the sequence of pages visited, and the delta sequence of timestamps from the current session passed as dynamic features, and, user behavior data from historical time period and user profile information as static features.

\textbf{Ablation study:} 
We have conducted an ablation study for both real-time churn and intended task detection. For ablation we have considered the following models: (a) LWM: only the long window model with automated clickstream features computed over the last few weeks; (b) SWM: only the short window model with clickstream in the last one hour from the ref-ts; (c) LSWCM: a joint neural network model with recent clicks over the short one hour window passing through the network defined in Eq.~\eqref{eq:bilstm}-\eqref{eq:swm_pred_layer} and, the LWM features being concatenated at the flattened layer and then passed through an FNN to get final prediction. We also compare the latency of these models to evaluate real-time performance.

\subsection{Results}
\label{subsec:results}
All the experiments have been conducted on one ml.m5.12xlarge AWS Sagemaker instance with 32 cores and the CPU is used for both training and inferences for the neural network models.

\textbf{Comparison with baselines in churn identification:} With 3.197 lift at the top decile (DL$1$), BBE-LSWCM is 60\% better than the batch baseline models (Table~\ref{tab:ricom_comp}), 35\% better than any of the in-session models, 50\% better than survival models and finally, 30\% better than the next best performing joint model DIN \cite{zhou2018deep}. A larger gap between BBE-LSWCM and other models can be observed at the second decile lift (DL$2$), which justifies the robustness of BBE-LSWCM as compared to the baselines considered here. Moreover, BBE-LSWCM can early detect the churners with an ATWC$1$ of 71 hours which is 8\% higher compared to the next best joint model and at least 40\% higher than any of the batch or in-session models. In terms of AUROC, BBE-LSWCM has outperformed all the baselines.
\begin{table}[h]
\centering
\resizebox{.99\columnwidth}{.17\textwidth}{%
\begin{tabular}{cccccc}
\toprule
\textbf{\small Model type} & \textbf{\small Model} & \textbf{\small DL$1$} & \textbf{\small DL$2$} &  \textbf{\small AUROC} & \textbf{\small ATWC$1$(hrs.)}\\ 
  \hline
\multirow{4}{*}{Batch} & LR & 1.207 & 1.046 & 0.515 & 32.67  \\ 
&  GBDT & 1.958 & 1.748 & 0.671 & 39.18 \\ 
&   RF & 1.587 & 1.472 & 0.639 & 31.85 \\
&   PLSTM & 1.506 & 1.732 & 0.610 & 41.40 \\\hdashline
\multirow{3}{*}{In-session} & GBDT & 1.019 & 1.001 & 0.503 & 27.81  \\
& LSTM & 1.529 & 1.081 & 0.529 & 18.64 \\
& emb-LSTM & 2.029 & 1.578 & 0.572 & 33.91 \\\hdashline
\multirow{2}{*}{Survival} & batch-RSF & 1.679 & 1.621 & 0.645 & 65.31 \\ 
& in-ses-DeepSurv & 1.010 & 1.000 & 0.498 & 24.55 \\\hdashline
\multirow{3}{*}{Joint} & CDMM & 1.146 & 1.031 & 0.508 & 14.94 \\
& xDeepFM & 1.437 & 1.299 & 0.607 & 58.92 \\
& DIN & 2.184 & 1.652 & 0.667 & 63.84 \\\hdashline
 Proposed  & \textbf{BBE-LSWCM} & \textbf{3.197} & \textbf{2.871} & \textbf{0.813} & \textbf{70.88} \\\hline
  Non-predictive & random & 1 & 1 & 0.50 & 28.72 \\
   \toprule
\end{tabular}%
}
\caption{\small Performance in real-time churn detection w.r.t. baselines.}\vspace{-5mm}
\label{tab:ricom_comp}
\end{table}

\textbf{Results from ablation study:} Table~\ref{tab:ablation_comp} showcases the importance of each BBE-LSWCM component through ablation study. In terms of all three metrics (DL$1$, DL$2$, and AUROC), the LWM (LightGBM with Count-Vec clickstream features) and SWM (BiLSTM with embedded clickstream) have better performance as compared to any of the batch and in-session models reported in Table~\ref{tab:ricom_comp}. This justifies the modeling choices for the BBE-LSWCM framework.
\begin{table}[h]
\centering
\resizebox{.98\columnwidth}{.07\textwidth}{%
\begin{tabular}{c|cccc|cc}
\toprule
& \multicolumn{4}{c|}{Real-time churn} & \multicolumn{2}{c}{Intended-task} \\\hline
\textbf{\small Model} & \textbf{\small DL$1$} & \textbf{\small DL$2$} &  \textbf{\small AUROC} &  \textbf{\small Latency} &  \textbf{\small ovr-AUROC} & \textbf{\small Latency}\\\hline
 LWM & 2.196 & 2.139 &  0.690 & -- & 0.683 & -- \\
 SWM & 2.277 & 1.897 & 0.591 & 0\% & 0.701 & 0\% \\
 LSWCM & 2.561  & 2.219 & 0.732 & 102\% & 0.749 & 112\% \\
 \textbf{BBE-LSWCM} & \textbf{3.197} & \textbf{2.871} &  \textbf{0.813} & \textbf{18\%} & \textbf{0.806} & \textbf{33\%} \\
   \toprule
\end{tabular}%
}
\caption{\small Ablation study for churn and intended-task detection.}\vspace{-4mm}
\label{tab:ablation_comp}
\end{table}

For the real-time churn detection problem, LWM has a higher AUROC of 0.69 as compared to 0.591 of SWM, indicating the importance of user behavior information beyond the current session. Though SWM is marginally better than LWM in DL$1$, the lift score of SWM drops sharply at the 2nd decile, proving the limitations of the in-session-only methods for real-time churn detection. By combining LWM and SWM, BBE-LSWCM retains the good properties of both models and achieves a 17\% relative improvement in AUROC. Recall that the LSWCM model is a single joint neural network trained on both short-window and long-window features. Though this model has the best performance among all the considered baselines (20\% lower DL$1$ as of BBE-LSWCM), the average inference latency\footnote{Latency has been reported as a relative increase from the in-session SWM model.  Note that, latency computation excludes the time required to transform the raw clickstream to LWM and SWM input vectors, where also BBE-LSWCM has an advantage as LWM featurization is not required for real-time inference.} of this model increases by 102\% from the in-session SWM model. As BBE-LSWCM only needs to process in-session clickstream data and LWM output to generate the prediction, the latency of BBE-LSWCM has increased by only 18\% as compared to SWM, while it has achieved 25\% relative improvement in AUROC.

For the intended-task detection, unlike the churn problem, SWM (ovr-AUROC 0.701) has better predictive accuracy as compared to LWM (ovr-AUROC 0.683), as in-session user clickstream has more context regarding the task the user is interested in. In this case, BBE-LSWCM has achieved a 12\% better AUROC while maintaining a low real-time latency (only a 33\% increase relative to SWM as opposed to 112\% for the joint model LSWCM).

\textbf{Hyper-parameter analysis:}
In this section, we study the impact of different hyper-parameter settings of BBE-LSWCM for the intended task detection problem. We consider the following set of hyper-parameters: (1) LWM: \textit{n\_estimator} - number of estimators (or number of trees, or maximum number of iterations), varying within [50, 100, 500, 1000], \textit{max\_bin} - maximum number of bins to group non-missing features, taking values in [20, 50, 100, 255]; (2) SWM: number of units in BiLSTM layer and dropout proportion varying within [0.05, 0.1, 0.25, 0.5] and [8, 16, 32, 64] respectively; (3) BBS: number of blocks in [1, 3, 6] and block size from [2, 4, 10].

\begin{figure}[h]
\centering
\begin{subfigure}{.5\columnwidth}
  \centering
  \includegraphics[trim={0.5cm 0.1cm 0.1cm 0.1cm}, width=.85\columnwidth, height=0.6\columnwidth, scale=0.5]{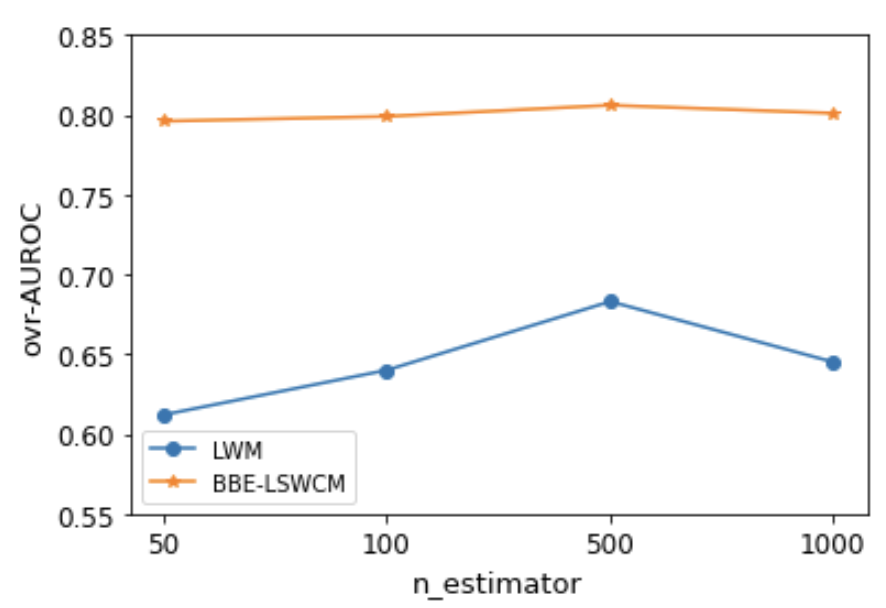}
\end{subfigure}%
\begin{subfigure}{.5\columnwidth}
  \centering
  \includegraphics[trim={0.1cm 0.1cm 0.5cm 0.1cm}, width=.85\columnwidth, height=0.6\columnwidth, scale=0.5]{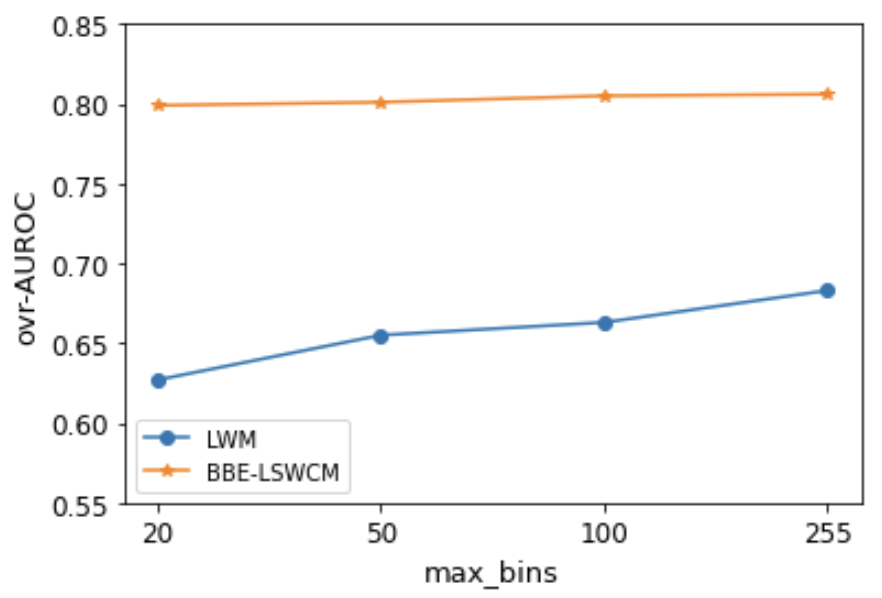}
\end{subfigure}\vspace{-1.5mm}
\caption{\small ovr-AUROC of BBE-LSWCM and LWM for varying LWM hyper-parameters}
\label{fig:hyper_param_lwm}\vspace{-2mm}
\end{figure}
As we can see in Fig.~\ref{fig:hyper_param_lwm}, the ovr-AUROC of LWM only model increases as the number of estimators increases till 500 and then it drops at 1000, whereas BBE-LSWCM is robust against the drop in performance of a single component (LWM in this case). In terms of the max\_bins, the performance of both LWM and BBE-LSWCM increases as we allow better distributional approximation through a higher number of bins to group the feature values.

\begin{figure}[h]
\centering
\begin{subfigure}{.5\columnwidth}
  \centering
  \includegraphics[trim={0.1cm 0.1cm 0.5cm 0.1cm}, width=.85\columnwidth, height=0.7\columnwidth, scale=0.5]{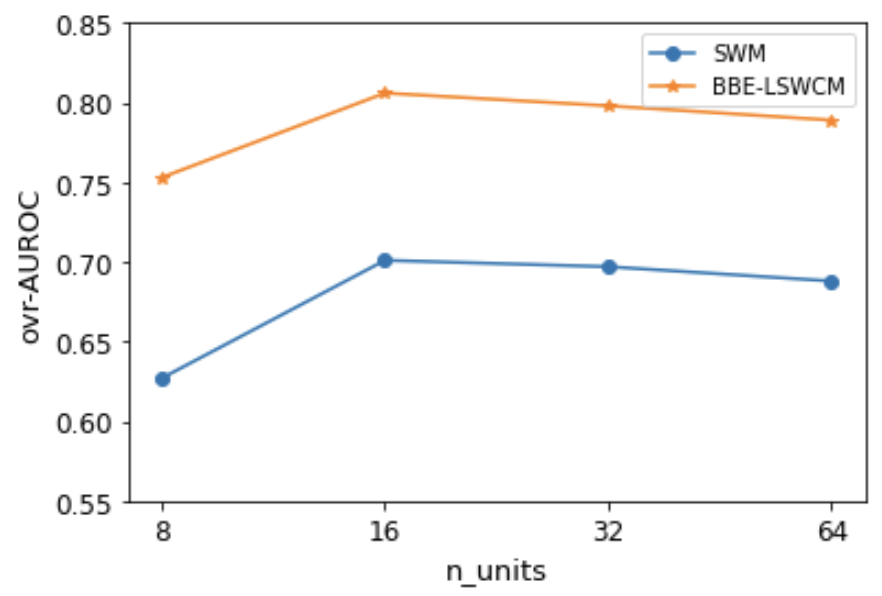}
\end{subfigure}%
\begin{subfigure}{.5\columnwidth}
  \centering
  \includegraphics[trim={0.1cm 0.1cm 0.5cm 0.1cm}, width=.85\columnwidth, height=0.7\columnwidth, scale=0.5]{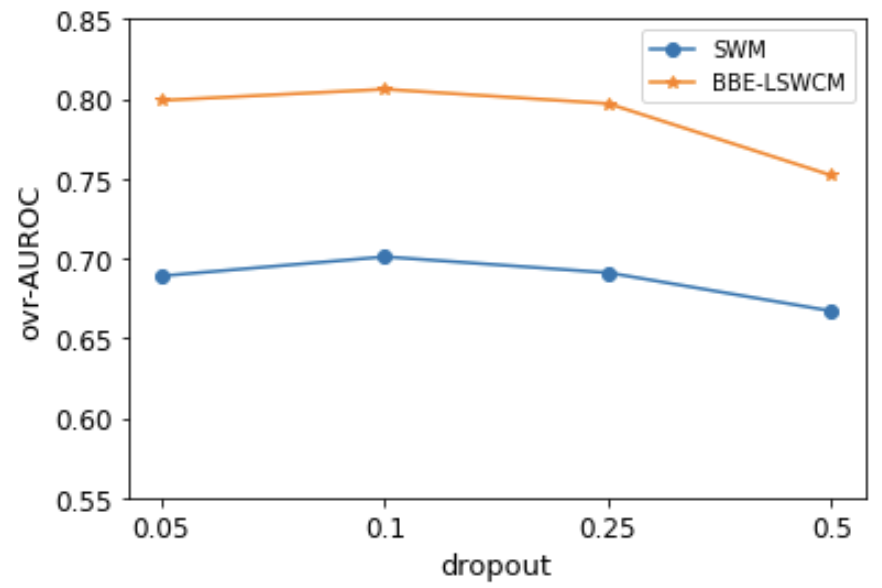}
\end{subfigure}\vspace{-1.5mm}
\caption{\small ovr-AUROC of BBE-LSWCM and SWM for varying SWM hyper-parameters}\vspace{-3mm}
\label{fig:hyper_param_swm}
\end{figure}
For varying SWM parameters (see Fig.~\ref{fig:hyper_param_swm}), the best accuracy has been achieved with number of units in BiLSTM as 16 and dropout proportion at 0.1. Note that, as SWM has more contribution to BBE-LSWCM for intended-task detection problem, the SWM hyper-parameters have a higher impact on BBE-LSWCM as compared to the LWM ones.

\begin{table}[h]
\small
\centering
\resizebox{.98\columnwidth}{.10\textwidth}
{%
    \begin{tabular}{cc|cc}
    \toprule
    \multicolumn{2}{c|}{\multirow{2}{*}{\textbf{Hyper-paramters}}} & \multicolumn{2}{c|}{relative to LSWCM w. random sampling} \\ \cline{3-4}
    & & \textbf{ovr-AUROC} & \textbf{Training Time}\\ 
      \hline
    n\_blocks = 1 & \multirow{3}{*}{block\_size = 4} & -1.290\% & -50.915\% \\ 
    n\_blocks = 3 &  & 7.610\% & -0.373\% \\ 
    n\_blocks = 6 & & 6.071\% & 36.387\% \\ \hdashline
    \multirow{3}{*}{n\_blocks = 3} & block\_size = 2 & 2.313\% & -28.981\% \\
    & block\_size = 4 & 7.610\% & -0.373\% \\ 
    & block\_size = 10 & 4.176\% & 58.549\% \\
       \toprule
    \end{tabular}%
}
\caption{\small Accuracy and training time comparison of BBE-LSWCM w.r.t. LSWCM for varying BBS hyper-parameters}
\label{tab:result_bbs}\vspace{-6mm}
\end{table}
As the BBS hyper-parameters decide the volume of the training data, we also report the impact of these hyper-parameters in the training time and accuracy as compared to fitting the LSWCM (the joint model) using a random sample from the whole data in Table~\ref{tab:result_bbs}. As the number of blocks (n\_blocks) and size of the blocks (block\_size) increases, the training time increases. When the number of blocks is 3 and block size is 4, i.e., per user around a maximum of 12 ref-ts are chosen through BBS, BBE-LSWCM has similar training time as of LSWCM (as the sample size is similar) while 7\% better ovr-AUROC. All the other hyper-parameters have been selected through 3-fold cross-validation on the training data.

\section{In-session Proactive Interventions}
\label{sec:model_prod}
\begin{figure}[h]
  \centering
  \includegraphics[trim={0.05cm 0.1cm 0.1cm 0.1cm}, width=.95\columnwidth, height=0.2\textheight, scale=0.9]{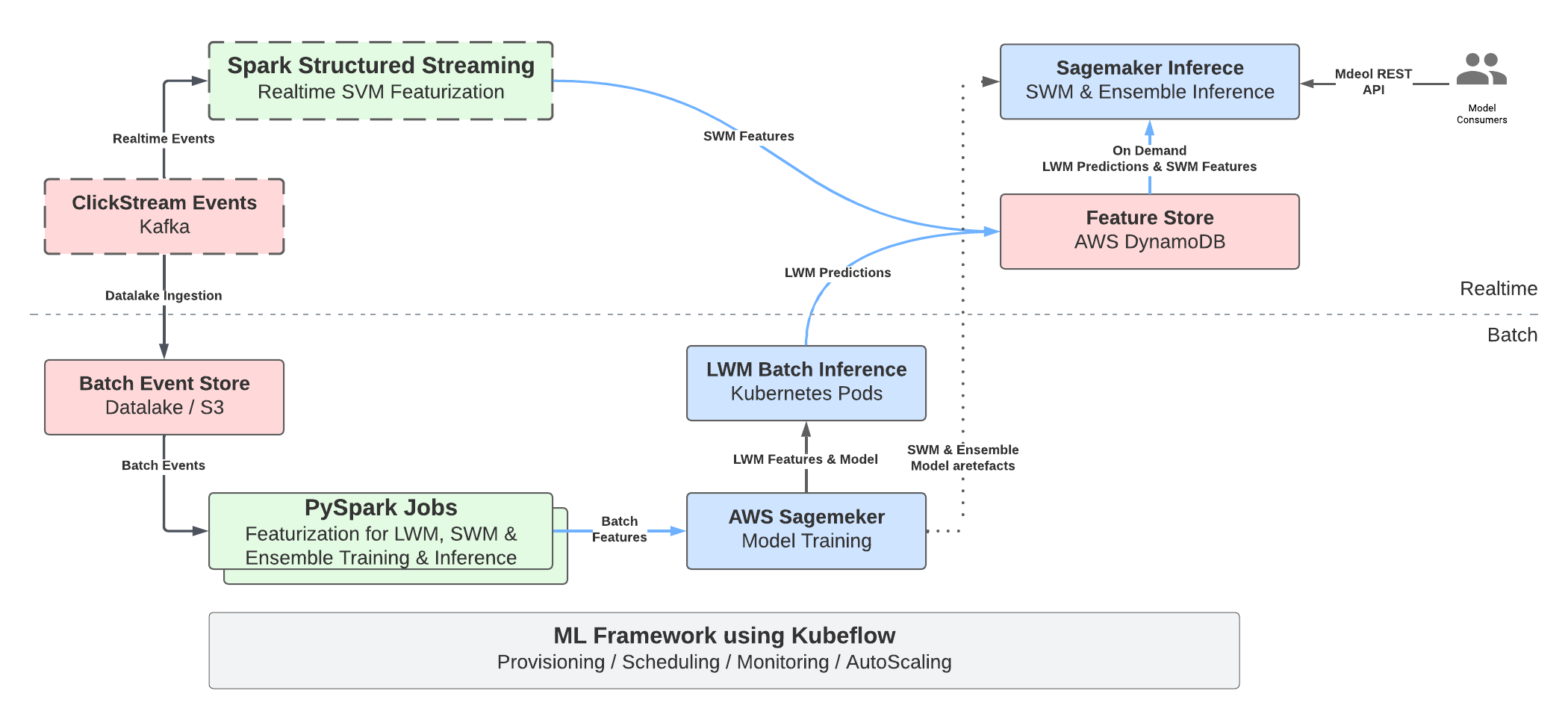}\vspace{-1mm}
  \caption{\small BBE-LSWCM production deployment architecture.}
    \label{fig:ricon_prod}\vspace{-4mm}
\end{figure}
BBE-LSWCM has been used to build a real-time prediction model for early detection of subscription cancellation in QBO. We have set up the data processing, model training and model inference as an end-to-end (e2e) automated pipeline (see Fig.~\ref{fig:ricon_prod}) with following components: (1) batch featurization using daily PySpark jobs scheduled on Kubernetes; (2) Real-time featurization using Spark’s Structured Streaming with executors running on Kubernetes; (3) AWS DynamoDB based low latency feature store; (4) AWS SageMaker jobs for LWM, SWM and EMM model training; (5) LWM parallel batch inference on Kubernetes pods; (6) AWS sagemaker inference for real-time SWM and EMM inference; (7) Kubeflow for orchestrating the entire pipeline.

This model has been integrated with the in-product intervention platform of QBO to run large-scale retention campaigns with real-time, contextual, and proactive interventions. An evaluation of the model (via an HTTP endpoint) at any given time for a customer provides the churn propensity of the user (High or Low with a probability score) based on the recent and long-window product usage in conjunction with user-profile information. When a model inference returns ‘High’ risk of churn, a proactive chat window pops up in the product with relevant help and the option to directly chat or call with a CRM agent. If the user accepts the proactive help, the intervention platform in that case evaluates the BBE-LSWCM model for intended task detection and passes the predicted intent of the user also to the CRM agent panel (Salesforce) to facilitate more contextual and personalized assistance.


\textbf{Online Results:} We have conducted an A/B test on the QBO first-time users for a complete month to track the effect of BBE-LSWCM-based real-time proactive interventions. We have observed a $\sim 30\%$ reduction in the FTU churn rate (p-value = 0.0057, power = 0.97) for the treatment group (intervention provided if detected as high risk) as compared to the control group (no intervention provided though detected as high risk). Moreover, the acceptance rate of the interventions at an aggregated user level is 4x as compared to other intervention campaigns which are powered by business rules as opposed to a machine learning system like BBE-LSWCM.

\section{Conclusion}
\label{sec:conclu}
We have proposed an efficient (cost-effective and low-latency) and robust (higher predictive accuracy compared to the baselines) clickstream modeling framework for real-time event prediction problems. By using a bootstrapped ensemble of short-window and long-window user behavior models, BBE-LSWCM has outperformed not only batch models but also other state-of-the-art in-session and survival models. We have also described how BBE-LSWCM has been used in a real industry setting for driving real-time proactive interventions for a large SaaS subscription commerce. One of the immediate future directions is to implement BBE-LSWCM for other real-time event prediction problems. In addition, though the local explainability of LWM can easily be generated through modules such as SHAP, LIME \cite{vstrumbelj2014explaining, ribeiro2016should}, generating explainability based on the embedded in-session clicks through the SWM and the EMM layer is another area to be explored.

\bibliographystyle{ACM-Reference-Format}
\bibliography{BBE-LSWCM}


\end{document}